\documentclass[runningheads]{llncs}
\usepackage{graphicx}
\usepackage{amsmath}
\usepackage{booktabs}
\usepackage{rotating}
\usepackage{makecell}
\usepackage{url}
\usepackage{xcolor}
\usepackage{multirow}

\begin{document}
\title{Walk this Way!}
\subtitle{Entity Walks and Property Walks for RDF2vec}
%
%
\author{Jan Portisch\inst{1,2}\orcidID{0000-0001-5420-0663} \and
Heiko Paulheim\inst{2}\orcidID{0000-0003-4386-8195}}
\authorrunning{J. Portisch and H. Paulheim}
%
\institute{SAP SE, Walldorf, Germany
\and
Data and Web Science Group, University of Mannheim, Germany\\
\email{\{jan,heiko\}@informatik.uni-mannheim.de}}
\maketitle              
\begin{abstract}
RDF2vec is a knowledge graph embedding mechanism which first extracts sequences from knowledge graphs by performing random walks, then feeds those into the word embedding algorithm word2vec for computing vector representations for entities. In this poster, we introduce two new flavors of walk extraction coined \emph{e-walks} and \emph{p-walks}, which put an emphasis on the structure or the neighborhood of an entity respectively, and thereby allow for creating embeddings which focus on similarity or relatedness. By combining the walk strategies with order-aware and classic RDF2vec, as well as CBOW and skip-gram word2vec embeddings, we conduct a preliminary evaluation with a total of 12 RDF2vec variants.

\keywords{RDF2vec \and Embedding \and Similarity \and Relatedness}
\end{abstract}

\section{Introduction}
RDF2vec~\cite{DBLP:conf/semweb/RistoskiP16} is an approach for embedding entities of a knowledge graph in a continuous vector space. It extracts sequences of entities from knowledge graphs, which are then fed into a word2vec encoder \cite{DBLP:conf/nips/MikolovSCCD13}. Such embeddings have been shown to be useful in downstream tasks which require numeric representations of entities and rely on a distance metric between entities that captures entity similarity and/or relatedness \cite{portisch2022knowledge}.

Different variants for walk extraction in RDF2vec have been proposed in the past, including the inclusion of weights in the random component \cite{DBLP:conf/wims/CochezRPP17} and the use of other walk strategies such as community hops and walklets \cite{DBLP:conf/dexaw/SteenwinckelVBW21}. Moreover, it has been shown recently that using an order-aware variant instead of classic word2vec improves the resulting embeddings~\cite{DBLP:conf/semweb/PortischP21}.

RDF2vec mixes the notion of similarity and relatedness. This can be seen, for example, in Table~\ref{tab:mannheim_most_similar}: The closest concepts in the vector space for \emph{Mannheim} are comprised of the city timeline, a person, the local ice hockey team, and close cities. All of these are \emph{related} to the city in a sense that they have a semantic relation to Mannheim (Peter Kurz, for instance, is Lord mayor of Mannheim). However, these concepts are not \emph{similar} to the city since a person and a city do not have much in common.

In this paper, we present two new variants of RDF2vec: \emph{p-RDF2vec} emphasizes structural properties of entities, i.e. their attributes, and consequently has a higher exposure towards similarity. \emph{e-RDF2vec} emphasizes the neighboring entities, i.e. the context of entities, and consequently has a higher exposure towards relatedness.

\begin{table}[t]
    \centering
    \begin{tabular}{r|l|l|l}
         \# & RDF2vec                 & p-RDF2vec          & e-RDF2vec                  \\   
         \hline   
        1	&	Ludwigshafen          & Arnsberg           & Ludwigshafen               \\
        2	&	Peter Kurz            & Frankfurt          & Timeline of Mannheim               \\
        3	&	Timeline of Mannheim  & Tehran             & Peter Kurz                          \\
        4	&	Karlsruhe             & Bochum             & Adler Mannheim                      \\
        5	&	Adler Mannheim        & Bremen             & Peter Kurze             \\
    \end{tabular}
    \caption{5 nearest neighbors to \emph{Mannheim} in RDF2vec (classic), p-RDF2vec, and e-RDF2vec trained on DBpedia (SG)}
    \label{tab:mannheim_most_similar}
\end{table}

\section{New Walk Flavors}
In the following, we define a knowledge graph $\mathcal{G}$ as a labeled directed graph $\mathcal{G} = (\mathcal{V},\mathcal{E})$, where $\mathcal{E} \subseteq \mathcal{V}\times\mathcal{R}\times\mathcal{V}$ for a set of relations $\mathcal{R}$. Vertices are subsequently also referred to as \emph{entities} and edges as \emph{predicates}.

Classic RDF2vec creates sequences of random walks. A random walk of length $n$ (for an even number n) for $w_0$ has the form
\begin{equation}
    w = (w_{-\frac{n}{2}}, w_{-\frac{n}{2}+1}, ..., w_0,..., w_{\frac{n}{2}-1}, w_\frac{n}{2})
\end{equation}
where $w_i \in \mathcal{V}$ if $i$ is even, and $w_i \in \mathcal{R}$ if $i$ is odd. For better readability, we stylize $w_i \in \mathcal{V}$ as $e_i$ and $w_i \in \mathcal{R}$ as $p_i$:
\begin{equation}
    w = (e_{-\frac{n}{2}}, p_{-\frac{n}{2}+1}, ..., e_0,..., p_{\frac{n}{2}-1}, e_\frac{n}{2})
\end{equation}
In the case of loops, it is possible that a walk contains an entity or edge more than once.

\begin{figure}[t]
    \centering
    \includegraphics[width=\textwidth]{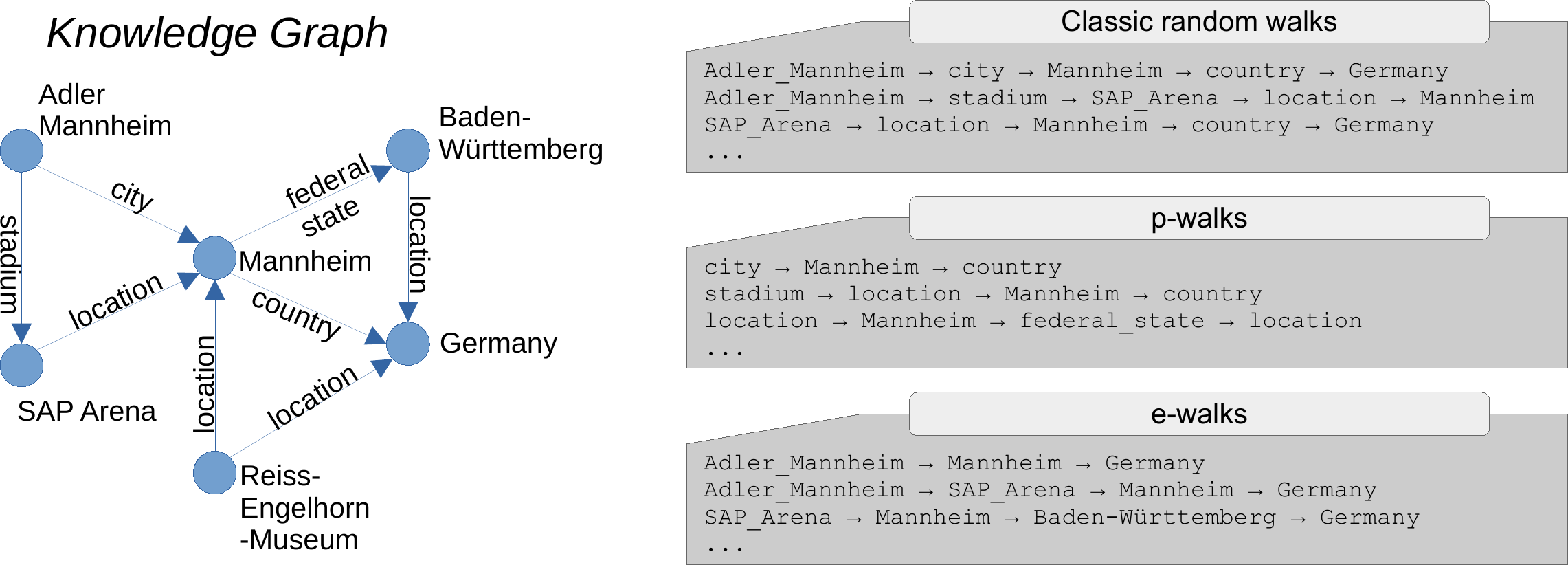}
    \caption{Illustration of the different walk types}
    \label{fig:schema}
\end{figure}

From the definition of random walks, we derive two other types of random walks (see Fig.~\ref{fig:schema}): A \emph{p-walk} $w_p$ is a subsequence of a walk $w$ which consists of only the focus entity $e_0$ and the predicates in the walk, i.e.,

\begin{equation}
w_p = (p_{-\frac{n}{2}+1}, p_{-\frac{n}{2}+3}, ..., e_0,..., p_{\frac{n}{2}-3}, p_{\frac{n}{2}-1})
\end{equation}

In contrast, an \emph{e-walk} consists only of the entities in the walk, i.e., 

\begin{equation}
w_e = (e_{-\frac{n}{2}}, e_{-\frac{n}{2}+2}, ..., e_0,..., e_{\frac{n}{2}-2}, e_{\frac{n}{2}})
\end{equation}

In other words: p-walks capture the \emph{structure} around an entity, while e-walks capture the \emph{context}. Thus, we hypothesize that embeddings computed from p-walks capture (structural) \emph{similarity}, while those computed from e-walks capture contextual similarity, which can also be understood as \emph{relatedness}.

\section{Evaluation}
We evaluate embeddings obtained using three different walk extraction strategies, i.e., classic walks, p-walks and e-walks, and training with classic word2vec as well as order-aware word2vec, using both the CBOW and skip-gram variants. This, in total, yields 12 different configurations for RDF2vec.\footnote{We generated 500 walks per node with a depth of 4, i.e., we perform 4 node hops. All embeddings are trained with a dimensionality of 200. The experiments were performed with jRDF2vec (\url{https://github.com/dwslab/jRDF2Vec}), which implements all the different variants used in this paper.} All embedding models are publicly available to download via KGvec2go~\cite{DBLP:conf/lrec/PortischHP20}.\footnote{\url{http://kgvec2go.org/download.html}}

For evaluation, we use the framework proposed in \cite{DBLP:conf/esws/PellegrinoAGRC20}, which consists of different tasks (classification, regression, clustering, analogy reasoning, entity relatedness, document similarity). We use a recent DBpedia release\footnote{\url{https://www.dbpedia.org/blog/snapshot-2021-09-release/}}.
The results are depicted in Table~\ref{tab:results}.
We can make a few interesting observations:
\begin{enumerate}
    \item In 12/20 cases, the best results are achieved with classic walks. p-walks yield the best results in 3/20 cases, e-walks do so in 5/20 cases.
    \item For entity relatedness, e-walks yield the best results, showing that those walks actually capture relatedness best.
    \item For document similarity, p-walks outperform the other approaches. One explanation could be that structural similarity of entities (e.g., politicians vs. athletes) is more important for that task.
    \item Semantic analogies are known to require both, relatedness and similarity.\footnote{For solving an analogy task like \emph{Paris is to France like Berlin is to X}, \emph{X} must be similar to \emph{France}, as well as related to \emph{Berlin}.} Therefore, one may expect both p-walks and e-walks to perform poorly which is indeed verified by our experiments.
    \item As observed in \cite{DBLP:conf/semweb/PortischP21}, the ordered variants almost always outperform the non-ordered ones, for all kinds of walks, except for the semantic analogy problems. This effect is even slightly stronger for p-walks and e-walks than for classic RDF2vec.
    \item Generally, skip-gram (and its ordered variant) are more likely to yield better results than CBOW.
\end{enumerate}
Table~\ref{tab:mannheim_most_similar} shows the five closest concepts for classic RDF2vec and the extensions presented in this paper. It can be seen that classic and e-RDF2vec have an exposure towards relatedness while p-RDF2vec results in similar entities (i.e., only cities) being retrieved.

\begin{sidewaystable}[t]\centering
\caption{Result of the 12 RDF2vec variants on 20 tasks. The best score for each task is printed in bold. The suffix $_{oa}$ marks the ordered variant of RDF2vec.}
\label{tab:results}
\footnotesize

\begin{tabular}{lll||cccc|cccc|cccc}
\toprule
\multicolumn{3}{c||}{ } & \multicolumn{4}{c|}{Classic RDF2vec} & \multicolumn{4}{c|}{e-RDF2vec} & \multicolumn{4}{c}{p-RDF2vec} \\
Task & Metric & Dataset & sg & sg$_{oa}$ & cbow & cbow$_{oa}$ & sg & sg$_{oa}$ & cbow & cbow$_{oa}$ & sg & sg$_{oa}$ & cbow & cbow$_{oa}$ \\ 
\toprule
Classification & ACC & AAUP & 0.706 & 0.713 & 0.643 & 0.690 & 0.696 & \textbf{0.717} & 0.703 & 0.690 & 0.564 & 0.623 & 0.551 & 0.612\\ 
 & & Cities & \textbf{0.818} & 0.803 & 0.725 & 0.723 & 0.770 & 0.743 & 0.750 & 0.702 & 0.606 & 0.677 & 0.501 & 0.707\\ 
 & & Forbes & \textbf{0.623} & 0.605 & 0.575 & 0.600 & 0.608 & 0.605 & 0.612 & 0.600 & 0.581 & 0.610 & 0.560 & 0.578\\ 
 & & Metacritic Albums & 0.586 & 0.585 & 0.536 & 0.532 & 0.596 & 0.583 & 0.564 & 0.584 & 0.634 & 0.632 & 0.569 & \textbf{0.667}\\ 
 & & Metacritic Movies & 0.726 & 0.716 & 0.549 & 0.626 & 0.724 & \textbf{0.732} & 0.686 & 0.676 & 0.610 & 0.660 & 0.535 & 0.663\\ 
\hline 
Clustering & ACC & \makecell[l]{Cities and\\Countries (2k)} & 0.789 & 0.900 & 0.520 & \textbf{0.917} & 0.726 & 0.726 & 0.668 & 0.660 & 0.605 & 0.520 & 0.637 & 0.733\\ 
 & & Cities and Countries & 0.587 & 0.760 & 0.783 & 0.720 & 0.749 & 0.766 & \textbf{0.820} & 0.745 & 0.687 & 0.782 & 0.787 & 0.728\\ 
 & & \makecell[l]{Cities, Albums\\Movies, AAUP,\\Forbes} & 0.829 & \textbf{0.854} & 0.547 & 0.652 & 0.759 & 0.828 & 0.557 & 0.719 & 0.598 & 0.798 & 0.663 & 0.748\\ 
 & & Teams & 0.909 & 0.931 & 0.940 & 0.925 & 0.889 & 0.926 & 0.916 & 0.931 & \textbf{0.941} & 0.938 & 0.940 & 0.580\\ 
\hline 
Regression & RMSE & AAUP & 65.985 & \textbf{63.814} & 77.250 & 66.473 & 67.337 & 65.429 & 70.482 & 69.292 & 80.318 & 72.610 & 96.248 & 77.895\\ 
 & & Cities & 15.375 & \textbf{12.782} & 18.963 & 19.287 & 17.017 & 16.913 & 17.290 & 20.798 & 20.322 & 17.214 & 24.743 & 20.334\\ 
 & & Forbes & 36.545 & \textbf{36.050} & 39.204 & 37.067 & 38.589 & 38.558 & 39.867 & 36.313 & 37.146 & 36.374 & 37.947 & 38.952\\ 
 & & Metacritic Albums & 15.288 & 15.903 & 15.812 & 15.705 & 15.573 & 15.785 & 15.574 & \textbf{14.640} & 15.178 & 14.869 & 15.000 & 16.679\\ 
 & & Metacritic Movies & \textbf{20.215} & 20.420 & 24.238 & 23.362 & 20.436 & 20.258 & 23.348 & 22.518 & 23.235 & 22.402 & 23.979 & 22.071\\ 
\hline 
\makecell[l]{Semantic\\Analogies} & ACC & \makecell[l]{capital\\country entities} & \textbf{0.957} & 0.864 & 0.810 & 0.789 & 0.794 & 0.747 & 0.660 & 0.397 & 0.008 & 0.091 & 0.000 & 0.036\\ 
 & & \makecell[l]{all capital\\country entities} & \textbf{0.905} & 0.857 & 0.594 & 0.758 & 0.657 & 0.591 & 0.359 & 0.592 & 0.014 & 0.073 & 0.002 & 0.052\\ 
 & & currency entities & \textbf{0.574} & 0.535 & 0.338 & 0.447 & 0.309 & 0.193 & 0.198 & 0.297 & 0.006 & 0.076 & 0.002 & 0.085\\ 
 & & city state entities & \textbf{0.609} & 0.578 & 0.507 & 0.442 & 0.459 & 0.484 & 0.250 & 0.361 & 0.009 & 0.048 & 0.000 & 0.036\\ 
\hline 
\makecell[l]{Entity\\Relatedness} & \makecell[l]{Kendall\\Tau} &  & 0.747 & 0.716 & 0.611 & 0.547 & \textbf{0.832} & 0.800 & 0.726 & 0.779 & 0.432 & 0.768 & 0.568 & 0.737\\ 
\hline 
\makecell[l]{Document\\Similarity} & \makecell[l]{Harmonic\\Mean} &  & 0.237 & 0.230 & 0.283 & 0.209 & 0.275 & 0.250 & 0.170 & 0.111 & 0.193 & \textbf{0.382} & 0.296 & 0.256\\ 
\bottomrule
\end{tabular}
\end{sidewaystable}

\section{Conclusion and Future Work}
In this work, we have shown that p-walks and e-walks are interesting alternatives, which, in particular in combination with the order-aware variant of RDF2vec, can outperform classic RDF2vec embeddings. Moreover, we have seen that using p-walks and e-walks can help create embeddings whose distance function reflects similarity and relatedness respectively.

At the same time, the evaluation is still not very conclusive. Therefore, we aim at compiling collections of synthetic test cases which will allow us to make clear statements about which techniques are promising for which kind of problem.

Another interesting avenue for future research is the combination of different embeddings. In cases where aspects of two or three different embedding techniques are relevant, those can be combined and fed into a downstream classification system.

\bibliographystyle{splncs04}
\bibliography{references.bib}

\end{document}